\documentclass[10pt,twocolumn,letterpaper]{article}

\usepackage{iccv}
\usepackage{times}
\usepackage{epsfig}
\usepackage{graphicx}
\usepackage{amsmath}
\usepackage{amssymb}
\usepackage{array}
\usepackage{iccv,times,graphicx,amsmath,amssymb,booktabs,tabulary,multirow,overpic,xcolor}

\newlength\savewidth\newcommand\shline{\noalign{\global\savewidth\arrayrulewidth
  \global\arrayrulewidth 1pt}\hline\noalign{\global\arrayrulewidth\savewidth}}

\usepackage[breaklinks=true,bookmarks=false]{hyperref}

\iccvfinalcopy 

\def\iccvPaperID{2259} 

\ificcvfinal\fi
\begin{document}

\title{InstaBoost: Boosting Instance Segmentation via Probability Map Guided Copy-Pasting}

\author{Hao-Shu Fang\footnotemark[1], Jianhua Sun\footnotemark[1], Runzhong Wang\footnotemark[1], Minghao Gou, Yong-Lu Li, Cewu Lu\footnotemark[4]\\
Shanghai Jiao Tong University, China\\
{\tt\small fhaoshu@gmail.com}
{\tt\small \{gothic,runzhong.wang,gmh2015,yonglu\_li,lucewu\}@sjtu.edu.cn}
}

\maketitle
\renewcommand{\thefootnote}{\fnsymbol{footnote}}
\footnotetext[1]{contributed equally to this paper}
\footnotetext[4]{Cewu Lu is the corresponding author: lucewu@sjtu.edu.cn. Cewu Lu is a member of MoE Key Lab of Artificial Intelligence, AI Institute, Shanghai Jiao Tong University, and SJTU-SenseTime AI lab.}

\begin{abstract}
Instance segmentation requires a large number of training samples to achieve satisfactory performance and benefits from proper data augmentation. To enlarge the training set and increase the diversity, previous methods have investigated using data annotation from other domain (e.g. bbox, point) in a weakly supervised mechanism. In this paper, we present a simple, efficient and effective method to augment the training set using the existing instance mask annotations. Exploiting the pixel redundancy of the background, we are able to improve the performance of Mask R-CNN  for \textbf{1.7 mAP} on COCO dataset and \textbf{3.3 mAP} on Pascal VOC dataset by simply introducing random jittering to objects. Furthermore, we propose a location probability map based approach to explore the feasible locations that objects can be placed based on local appearance similarity. With the guidance of such map, we boost the performance of R101-Mask R-CNN on instance segmentation from \textbf{35.7 mAP to 37.9 mAP} without modifying the backbone or network structure. Our method is simple to implement and does not increase the computational complexity. It can be integrated into the training pipeline of any instance segmentation model without affecting the training and inference efficiency. Our code and models have been released at \hyperlink{https://github.com/GothicAi/InstaBoost}{https://github.com/GothicAi/InstaBoost}.
\end{abstract}

\section{Introduction}
Instance segmentation aims to simultaneously perform instance localization and classification and outputs pixel-level masks denoting the detected instance. It plays an vital role in computer vision and has many practical applications in autonomous driving~\cite{Cordts2016Cityscapes}, robotic manipulation~\cite{richtsfeld2012segmentation}, HOI detection~\cite{li2019transferable, qi2018learning} etc. Recent researches have proposed effective CNN (Convolution Neural Networks) architectures ~\cite{liCVPR17fcis,heICCV17maskrcnn} for the problem. To fully exploit the power of CNN, a large number of training data is indispensable. However, obtaining the annotations of pixel-wise masks is labor intensive, and thus limits the number of available training samples.

To tackle this problem, previous works utilize the data from other domains and conducted weakly supervised learning to obtain extra information. These researches mainly follow two lines: i) transform annotations from other domain to object masks~\cite{dai2015boxsup, lin2016scribblesup} or ii) utilize data from other domain as extra regularization term~\cite{gong2017look,bearman2016point}. However, few of these works investigate leveraging the existing mask annotations to augment the training set.

\begin{figure}[t]
\begin{center}
   \includegraphics[width=1.0\linewidth]{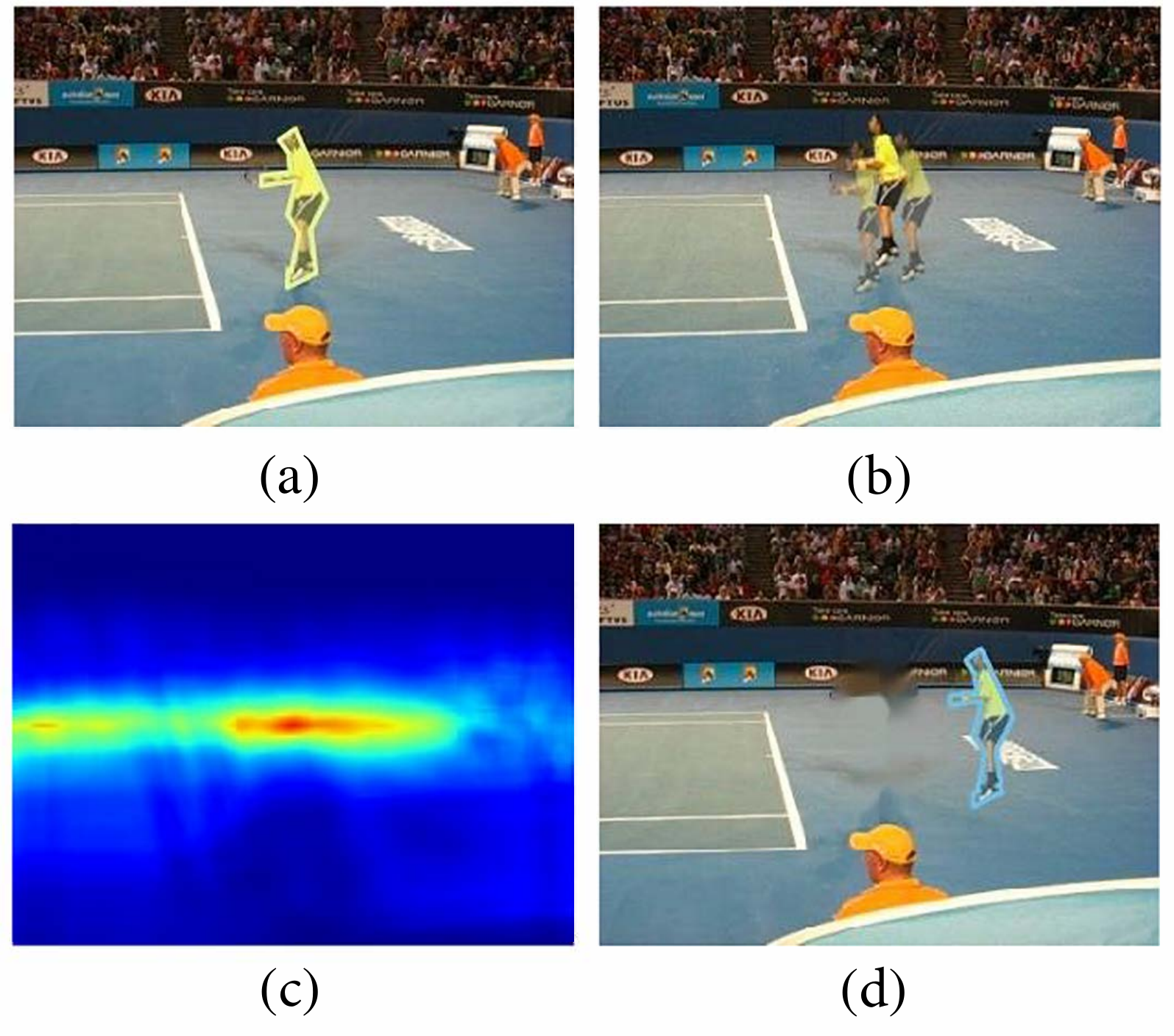}
\end{center}
\vspace{-10pt}
   \caption{An example of random InstaBoost and appearance consistency heatmap guided InstaBoost. (a) An original image with ground truth mask label from COCO dataset. (b) The result of random InstaBoost. Multiple pastes are visualized showing the randomness. (c) Appearance consistency heatmap of this image. (d) The result of appearance consistency heatmap guided InstaBoost.}
\label{fig:introduction}
\vspace{-10pt}
\end{figure}

Recently, crop-and-paste data augmentation has been exploited in the area of instance detection~\cite{dwibediICCV17cut} and object detection~\cite{dvornikECCV18modeling}. They crop the object using their masks and paste them on a random chose background randomly or according to the visual context. However, these data augmentation method does not work in the area of instance segmentation, as dataset priors are not efficiently exploited, resulting in poor performance in our experiments. Meanwhile, adopting a deep context model~\cite{dvornikECCV18modeling} introduces significant computational overhead, making it less practical in real-world applications.

In this paper, we first propose a simple but surprisingly effective random augmentation technique. Inspired by the stochastic grammar of images~\cite{zhu2007stochastic}, we paste objects in the neighboring of its original position, with additional small jittering on scale and rotation. Namely \emph{random InstaBoost}, such method brings  1.7 mAP improvement with Mask R-CNN on COCO instance segmentation benchmark.

Further, we look back to the area of visual perception, from which we get inspiration for a better-refined position transformation scheme. Previous research in Bayesian approaches to brain function shows the brain's ability to extract perceptual information from sensory data was modeled in terms of probabilistic estimation~\cite{wade2013visual} and visual inference requires prior experience of the world~\cite{arnheim1969visual}. These researches shed light on the area of crop-paste data augmentation for instance segmentation.

Intuitively, there exists a probability map representing reasonable placement that aligns with real-world experience. Inspired by~\cite{field1993contour}, we link such probability map to appearance consistency heatmap, which is based on local contour similarity since the background usually has redundancy in continuous, but non-aligned features. We sample feasible locations from the heatmap and conduct crop-paste data augmentation, reaching in total \textbf{2.2 mAP} improvement on the COCO dataset. Such a scheme is denoted as \emph{appearance consistency heatmap guided InstaBoost}. An example of our appearance consistency heatmap is shown in Fig.\ref{fig:introduction}.

We conduct exhaustive experiments on the Pascal VOC dataset and COCO dataset. By augmenting through appearance consistency heatmap guided InstaBoost, we are able to achieve  2.2 mAP improvement of COCO instance segmentation and 3.9 mAP on Pascal dataset.


\section{Related work}

\noindent \textbf{Instance mask segmentation.} Combining instance detection and semantic segmentation, instance segmentation~\cite{DaiCVPR16inst, heICCV17maskrcnn, lin2014complex, wang2017saliency, liCVPR17fcis, wang2019salient, xu2019ese, lu2019see} is a much harder problem. Earlier methods either propose segmentation candidates followed by classification~\cite{pinheiro2015learning}, or associate pixels on the semantic segmentation map into different instances~\cite{bleau2000watershed}.
Recently, FCIS~\cite{liCVPR17fcis} proposed the first fully convolutional end-to-end solution to instance segmentation, which predicted position-sensitive channels~\cite{daiECCV16instance} for instance segmentation. This idea is further developed by~\cite{chenCVPR18masklab} which outperforms competing methods on the COCO dataset~\cite{linECCV14coco}. With the help of FPN~\cite{linCVPR17fpn} and a precise pooling scheme named \emph{RoI Align}, He \etal~\cite{heICCV17maskrcnn} proposed a two-step model Mask R-CNN that extends Faster R-CNN framework with a mask head and achieves state-of-the-art on instance segmentation~\cite{MSCOCO} and pose estimation~\cite{li2019crowdpose} tasks.
Although these methods have reached impressive performance on public datasets, those heavy deep models are hungry for an extremely large number of training data, which is usually not available in real-world applications. Furthermore, the potential of large datasets are not fully exploited by existing training methods.

\noindent \textbf{Instance-level augmentation.} One branch of recent work has emerged with more precise instance-level image augmentation, laying potential to fully exploit the supervised information in the existing dataset~\cite{dwibediICCV17cut, khoreva2018lucid, daiArxiv18learning, dvornikECCV18modeling, fang2018wshp, huang2018cooperative, xu2018srda}. Dwibedi~\etal~\cite{dwibediICCV17cut} improved instance detection by simple cut-and-paste strategy with extra instances that have annotated masks.
Khoreva~\etal~\cite{khoreva2018lucid} generate pairs of synthetic images for video object segmentation using cut-and-paste method. However, the object position is uniformly sampled and they just need to guarantee that changes between image pairs are kept small. Such setting does not work for image-level instance segmentation, as we demonstrated in our experiments that randomly pasted object will decrease the segmentation accuracy. Another recent work~\cite{dvornikECCV18modeling} proposed a context model to place segmented objects at backgrounds with proper context and demonstrated that it can improve objection detection on the Pascal VOC dataset. Such method requires training an extra model and preprocessing data offline. In this paper, we propose a simple but effective online augmentation method, which is the first attempt that successfully improve overall accuracy on COCO ~\textbf{instance segmentation}, as to the best of our knowledge.

\section{Our approach}

\subsection{Overview}
Given a cropped object patch from a specific image, the placement of that patch on the image can be defined by the affine transformation matrix
\begin{equation}
    \mathbf{H}=
    \left[ \begin{array}{ccc}
         s \cos{r} & s \sin{r} & t_x \\
         - s \sin{r} & s \cos{r} & t_y \\
         0 & 0 & 1
    \end{array} \right]
    \label{eq:aff_matrix}
\end{equation}
where $t_x$, $t_y$ denote the coordinate shift in $x,y$-axis respectively, $s$ denotes the scale variance and $r$ denotes the rotation in degrees. Thus, the placement can be uniquely determined by a 4D tuple
\begin{equation}
    \mathcal{B} = \left\{(t_x,t_y,s,r)\right\} \quad t_x, t_y, r \in \mathbb{R}, s \in \mathbb{R}^+
    \label{eq:trans_tuple}
\end{equation}

From the view of stochastic grammar of images~\cite{zhu2007stochastic}, a probabilistic model can be defined on this 4D space to learn the natural occurrence frequency of objects and then sampled to synthesize a large number of configurations to cover novel instances in the test set.
By this end, we define probability density function $f(\cdot)$ measuring how reasonable it is to paste the object $O$ on the given image $I$, following a specific transformation tuple. Assuming $(x_0, y_0)$ as the object's original coordinate and $x=x_0+t_x, y=y_0+t_y$ are new coordinates, a probability map $P$ is defined on set $\mathcal{B}$, which is given as
\begin{equation}
    P(x, y, s, r\,|\,I,O) = f(t_x,t_y,s,r\,|\,I,O).
\label{eq:p_xysr}
\end{equation}
the given image and object conditions $(I, O)$ will be omitted for simplicity in the following context. Specifically, the identity transform $(x_0, y_0, 1, 0)$ which corresponds to the original paste configuration should have the highest probability, i.e.
\begin{equation}
    \arg \max P(x, y, s, r) = (x_0, y_0, 1, 0)
\end{equation}

Intuitively, in a small neighbor area of $(x_0, y_0, 1, 0)$, our probability map $P(x, y, s, r)$ shall also be high-valued since images are usually continuous and redundant in pixel level. Based on such observation, we propose a simple but effective augmentation approach: object jittering that randomly samples transformation tuples from the neighboring space of identity transform $(x_0, y_0, 1, 0)$ and paste the cropped object following affine transform $\mathbf{H}$. Experimental result in Sec.~\ref{sec:exp_inst_seg} shows the surprising effectiveness of  this simple data augmentation strategy.

In addition, inspired by \cite{arnheim1969visual}, the feasible location of  $(x, y)$ can be further extended without being restricted to the neighboring area of $(x_0, y_0)$ if the background shares a similar pattern for a wide range. Therefore, we proposed a simple appearance consistency heatmap to utilize the redundancy in continuous, but non-aligned features of background. With the guidance of such heatmap, we can maximize the utility of our object jittering.

In Sec.~\ref{sec:rand_instaboost}, we introduce the pipeline of our vanilla object jittering, while the generation and adoption of our appearance consistency map will be detailed in Sec.~\ref{sec:heatmap}.

\subsection{Random InstaBoost}
\label{sec:rand_instaboost}

A simple but effective augmentation approach named random InstaBoost is proposed, which draws a sample from an instance segmentation dataset, separate its foreground and background with ground truth annotations aided with matting and inpainting, and apply a restricted random transform to generate an augmented image. With visually appealing images generated via InstaBoost, experiments show the effectiveness of random InstaBoost, achieving 1.7 mAP improvement on COCO instance segmentation. Random InstaBoost mainly contains two steps: i) instance and background preparation via matting and inpainting and ii) random transform sampled from neighboring space of identity transform.

\begin{figure}
    \centering
    \includegraphics[width=0.46\textwidth]{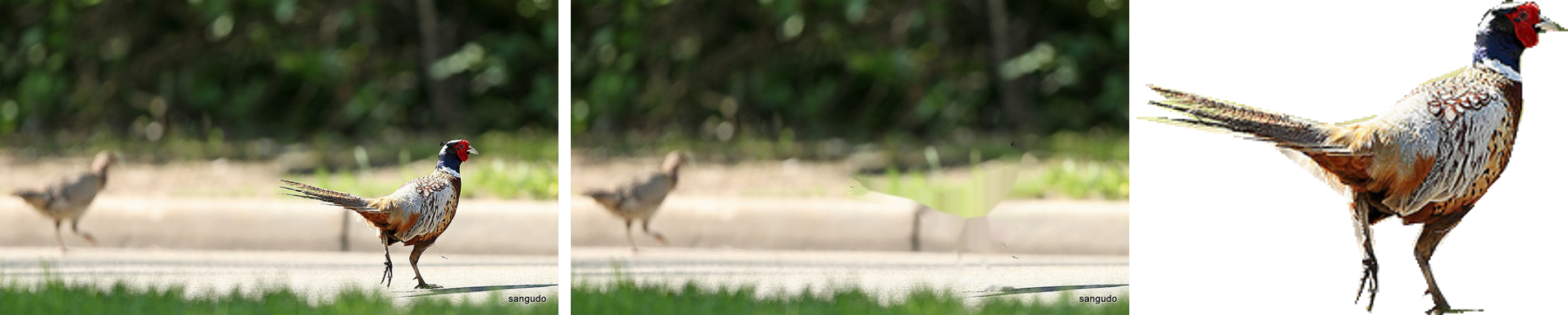}
    \caption{Example for inpainting and matting visualization. From left to right is original image, inpainting result and instance obtained by matting.}
    \label{fig:imgprocess}
\vspace{-10pt}
\end{figure}

\begin{figure*}[tb!]
\centering
\includegraphics[width=0.9\textwidth]{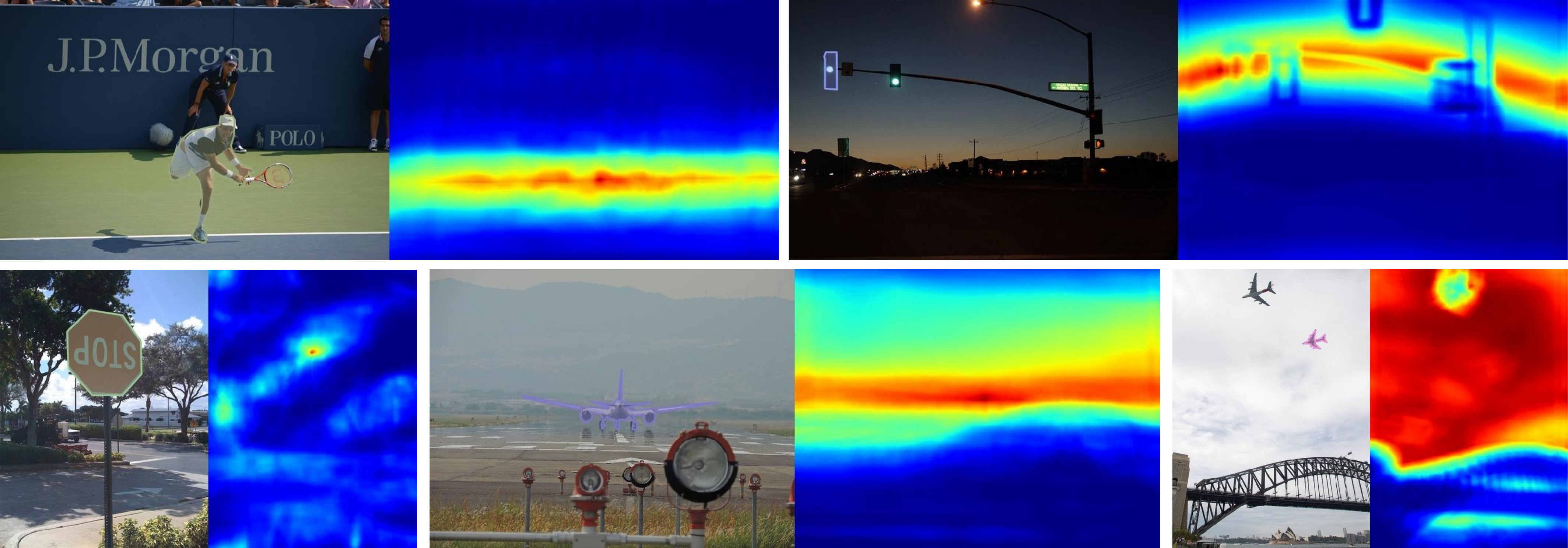}
\caption{Examples of appearance consistency heatmap. The left part of each example is the original image with an instance and the right part is the appearance consistency heatmap for that image. The red region is of high appearance consistency while the blue region is of low appearance consistency.}
\label{fig:2tuple}
\vspace{-10pt}
\end{figure*}

\noindent \textbf{Instance and background preparation.} Given an image with ground truth labels for instance segmentation, we need to separate the target instance and the background,
where the annotation of an instance segmentation dataset has already given sufficient information. However, in popular datasets e.g. COCO~\cite{linECCV14coco}, annotations are stored in the format of boundary points and edges, leading to a disappointing situation where the outline is zigzag. To overcome such issue, matting~\cite{he2011global} is adopted to get a smoother outline with the alpha channel, which is much more similar to the actual situation. In such a manner, instances can be cut off from the original image properly.

After the cutting step, we get a reasonable instance patch and an incomplete background with an instance-shaped hole on it.
Inpainting method~\cite{bertalmioCVPR01navier} are adopted to fill in such holes. Fig. \ref{fig:imgprocess} shows an example for inpainting and matting visualization.

\noindent \textbf{Random transformation} With 4D tuple transformation parameters defined in Eq.~(\ref{eq:trans_tuple}), our simple but effective InstaBoost technique is proposed, where $(t_x, t_y, s, r)$ are all random variables sampled from uniform distribution in the neighboring space of identity transform $(0, 0, 1, 0)$. Slight blurring is introduced to the original image, which will not strongly violate the visual content in the original image, but parallelly provides additional supervision to train instance segmentation models.

\subsection{Appearance consistency heatmap guided InstaBoost}
\label{sec:heatmap}
The feasible transformation of $(x, y)$ coordinates is restricted in the neighborhood of $(x_0, y_0)$ in random InstaBoost, whose performance could be further elevated with a more complicated metric on the image, i.e. \emph{appearance consistency heatmap}, to better refine the position where the new instance is pasted. Regarded as one implementation of the probability metric in Eq.~(\ref{eq:p_xysr}), appearance consistency heatmap evaluates similarity on the RGB space, between any transformation $(x, y)$ with respect to $(x_0, y_0)$. Examples of appearance consistency heatmap on COCO~\cite{linECCV14coco} dataset are shown in Fig.~\ref{fig:2tuple}. Each example in Fig.~\ref{fig:2tuple} consists of two images, the left image is the original image from COCO dataset and the right one is the corresponding appearance consistency heatmap.

We derive $f(\cdot)$ in Eq.~(\ref{eq:p_xysr}) as three conditional probability functions $f_{xy}(\cdot)$, $f_{s}(\cdot)$ and $f_{r}(\cdot)$ denoting probability density function w.r.t. $(t_x,t_y)$ and $(s,r)$, respectively, whereby the formulation is simplified assuming the independence between $(t_x, t_y), s$ and $r$:
\begin{equation}
\begin{split}
    P(x, y, s, r) =& f_{xy}(t_x,t_y\, |\, s,r)f_{s}(s\,|\, r)f_r(r) \\
    =& f_{xy}(t_x, t_y)f_{s}(s)f_r(r)
\end{split}
\end{equation}
where $f_{s}(s), f_{r}(r)$ are uniform distributions adopted by random InstaBoost in Sec.~\ref{sec:rand_instaboost}. Appearance consistency heatmap $M$ is defined as the expectation of probability map $P$, given $x, y$, input image $I$ and object patch $O$, which is proportional to $f_{xy}(t_x, t_y)$
\begin{equation}
    M(x, y) = E\left[P(x, y)\right] \propto f_{xy}(t_x, t_y)
    \label{eq:heatmap}
\end{equation}
Details of the appearance consistency map will be given as follows.

\begin{figure}[tb!]
\centering
\includegraphics[width=0.45 \textwidth]{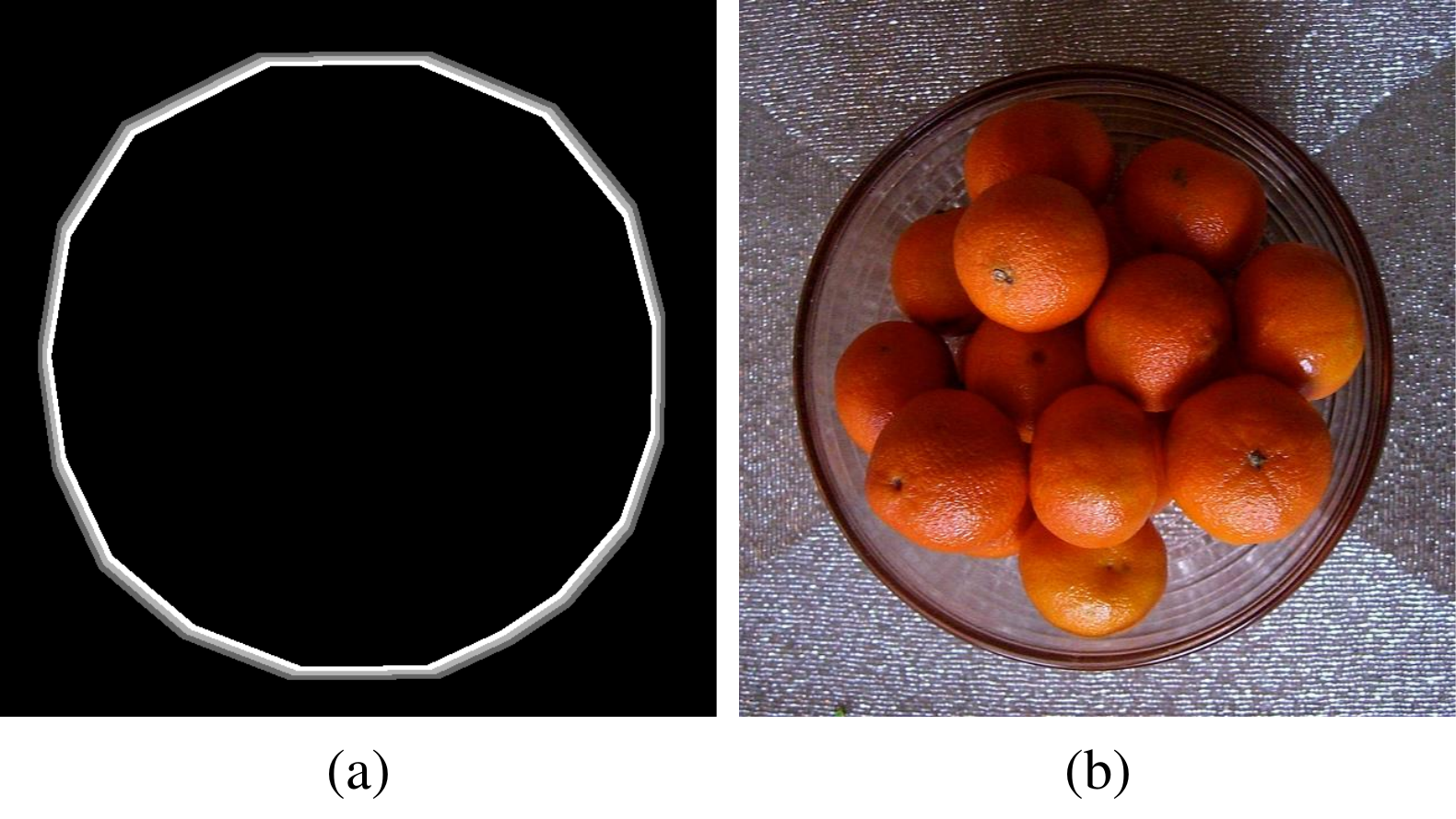}
\caption{One example of contour areas of appearance consistency heatmap. (a) The effective contour area of this image. (b) The original image.}
\label{fig:trimap}
\vspace{-10pt}
\end{figure}

\subsubsection{Appearance consistency heatmap}

\noindent \textbf{Appearance descriptor}. To measure the appearance similarity of an object patch pasted on two locations, we first need to define a descriptor which encodes the texture of the background in the neighbor area of the object. Intuitively, the influence of the ambient environment of the target instance on appearance consistency decreases with the increase of distance.

Based on this assumption, we define the appearance descriptor $\mathcal{D}(\cdot)$ as the weighted combination of three fixed width contour areas with different scales, which can be formulated as
\begin{equation}
    \mathcal{D}(c_x, c_y) = \{(\mathcal{C}_i(c_x,c_y), w_i) \,|\,i\in\{1,2,3\}\}
\end{equation}
where $\mathcal{C}_i$ denotes the contour area $i$ with weight $w_i$, given $c_x, c_y$ as the center of the instance. With $i=1$ being the most inside contour, we define $w_1 > w_2 > w_3$ emphasizing stronger consistency around neighboring areas of the original object. Fig.~\ref{fig:trimap} shows an example of contour areas of appearance consistency heatmap.

\begin{figure*}[tb!]
\centering
\includegraphics[width=\textwidth]{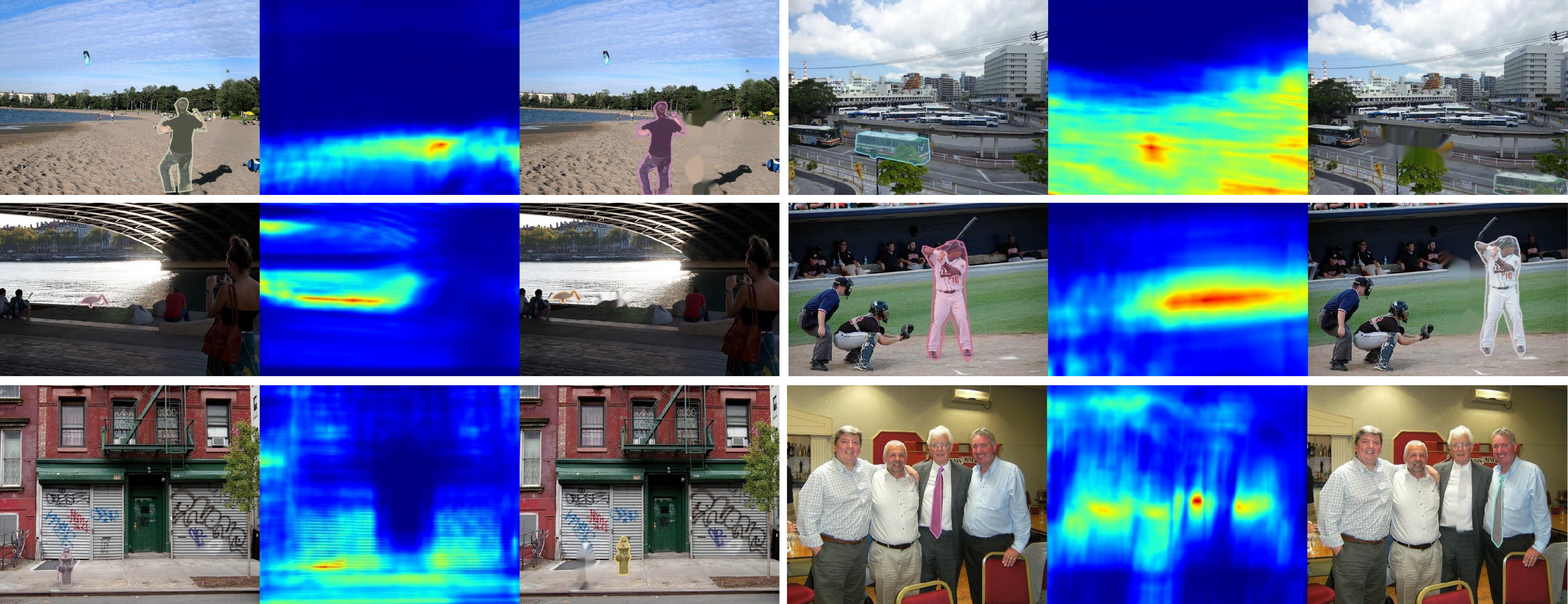}
\caption{Examples of appearance consistency heatmap guided InstaBoost. Each example consisted of the original image with an instance, appearance consistency heatmap and processed image from left to right.}
\label{fig:3tuple}
\vspace{-10pt}
\end{figure*}

\noindent \textbf{Appearance distance.} In this part, appearance distance is defined as local appearance consistency metric between pairs of appearance descriptor, i.e. instance centers. Since we have already defined affinity descriptor with three contour areas and corresponding weights, appearance distance between $\mathcal{D}_1 = \mathcal{D}(c_{1x}, c_{1y}), \mathcal{D}_2 = \mathcal{D}(c_{2x}, c_{2y})$ is defined as
\begin{equation}
    d(\mathcal{D}_1, \mathcal{D}_2) = \sum_{i=1}^3 \sum_{\substack{(x_1,y_1) \in \mathcal{C}_{1i} \\ (x_2,y_2) \in \mathcal{C}_{2i}}} w_i \Delta(I_1(x_1, y_1), I_2(x_2, y_2))
    \label{eqn:sim}
\end{equation}
where $(w_i, \mathcal{C}_{1i}) \in \mathcal{D}_1, (w_i, \mathcal{C}_{2i}) \in \mathcal{D}_2$. $I_k(x, y)$ denotes the RGB value of image $k$ on $(x,y)$ pixel coordinate. $\Delta$ can take any distance metric, where Euclidean distance is adopted in our implementation.

There occurs an exception that when part of the semantic consistency effective area locates outside of the background. For this situation, we consider the semantic consistency distance of this pixel equals to infinity (and therefore ignored).

\noindent \textbf{Heatmap generation.} By fixing $\mathcal{D}_0$ to the object's original position and scanning appearance distance $d(\mathcal{D}, \mathcal{D}_0)$ on all feasible $\mathcal{D}$ in the image, a heatmap is produced w.r.t. the center positions are taken by $\mathcal{D}$. Appearance distances are normalized and scaled via negative $\log$ for the heatmap $H$. The mapping is formulated as
\begin{equation}
    h(x) = -\log \left(\frac{x-m}{M-m}\right)
\end{equation}
where $M = \max \left(d(\mathcal{D}, \mathcal{D}_0)\right)$ represents the maximum distance in all candidate centers, $m = \min \left(d(\mathcal{D}, \mathcal{D}_0)\right)$ represents the minimum distance. Heatmap $H$ is generated with $h(\cdot)$ applied to every pixel in the background image, with respect to original instance's position $(x_0, y_0)$.

\subsubsection{Heatmap to transformation tuple}
\label{sec:heatmap2pos}
\noindent \textbf{Coordinate shift}. Transformation is performed according to a 4D tuple as introduced in Eq.~(\ref{eq:aff_matrix}, \ref{eq:trans_tuple}). As suggested in Eq.~(\ref{eq:heatmap}), heatmap values are proportional to the probability density function on $x, y$-axis, namely $f_{xy}(\cdot)$. Therefore, values in the appearance consistency heatmap are normalized and treated as probabilities, from which candidate points are sampled via Monte Carlo method. Compared to randomly sampling $(x,y)$ from the uniform distribution, the feasible area to placing the new object grows significantly, while avoiding pasting the instance onto semantically inconsistent backgrounds. Such operation on the heatmap introduces extra information for model training, which is an appealing feature for data augmentation.

\noindent \textbf{Scaling and rotation}. Scale and rotation parameters $(s, r)$ are sampled independently from uniform distribution in the neighboring of $(1, 0)$, as we assume independence among $(x, y), s, r$. Such practice is identical to our implementation of random InstaBoost in Sec.~\ref{sec:rand_instaboost}.

\subsubsection{Acceleration}
Following the steps described in Section~\ref{sec:heatmap}, we can successfully generate a heatmap for any target instance. However, computing the feature map is computationally inefficient as it needs to compute $W\times H$ semantic consistency distances for each point in the original effective area, where $W$ represents the width of the image and $H$ represents the height.
The time complexity comes to $\mathcal{O}(W^2 H^2)$ for computing Eq.~\ref{eqn:sim}, which is unacceptable in real-world applications. Therefore, we calculate the similarity map after resizing the original images to a fixed size and then upsample the heatmap to the original image size through interpolation. With such an acceleration strategy, appearance consistency heatmap is calculated in high quality and high speed, which is decisive in the implementation of our online InstaBoost algorithm.

\subsection{Training}
Our InstaBoost data augmentation strategy can be integrated into the training pipeline of any existing CNN based framework. During the training phase, the dataloader takes an image and applies InstaBoost strategy with a given probability, together with other data augmentation strategies. Our implementation of InstaBoost only introduces little CPU overhead to the original framework, together with parallel processing of dataloader that guarantees the efficiency during the training phase.

\begin{figure*}[tb!]
\centering
\includegraphics[width=\textwidth]{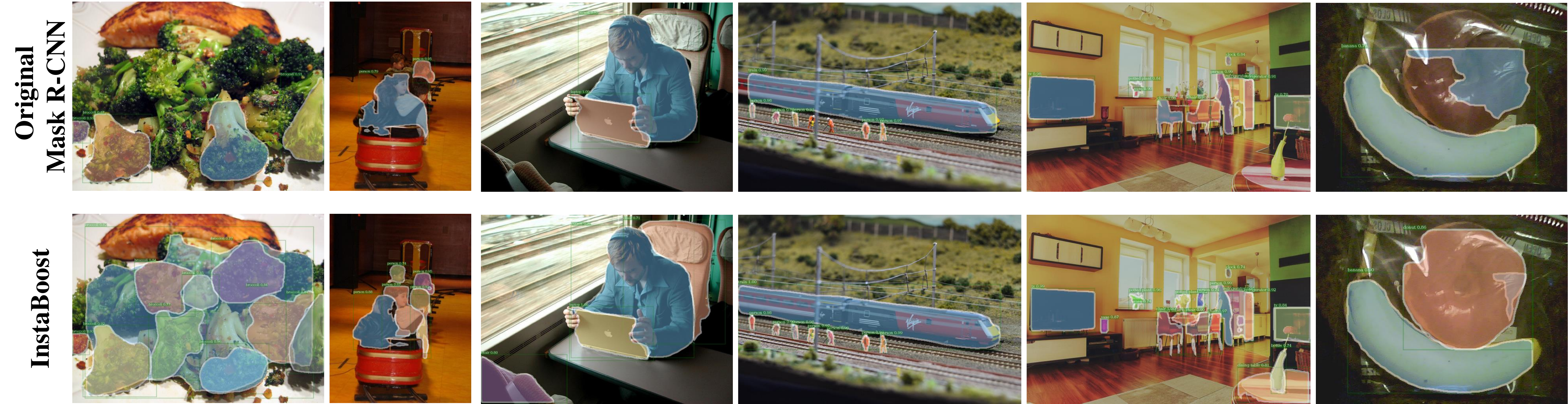}
\caption{Instance segmentation result of vanilla Mask R-CNN~\cite{heICCV17maskrcnn} (top) \emph{vs.} Mask R-CNN trained with InstaBoost (bottom). InstaBoost guarantees finer instance segmentation result.}
\label{fig:overview}
\vspace{-10pt}
\end{figure*}

\subsection{Discussion}
Previous method~\cite{dvornikECCV18modeling} investigated applying context model to explicitly model the consistency of the object and background in semantic space. Different from their approach, our appearance consistency map does not consider the semantic consistency explicitly but enforces the object to be pasted at places with similar background pattern on the original image. With such tight constraint, although some configurations that are semantic consistent but present a different background pattern may be pruned, we can guarantee that the generated images are visually coherent in most cases. Compared to~\cite{dvornikECCV18modeling}, our method can generate images that is more photorealistic and displays less blending artifacts, therefore introducing less noise when training the neural networks. Experimental results (Sec.~\ref{sec:exp_inst_seg}, Sec.~\ref{sec:analysis}) show the superior performance of our method in both qualitative and quantitative manner, while having a much more efficient implementation.

\begin{table*}[tb!]
\begin{center}
 \begin{tabular}{c|c|ccc|ccc}
   & Method & $AP^{det}$ & $AP^{det}_{50}$ & $AP^{det}_{75}$ & $AP^{det}_{S}$ & $AP^{det}_{M}$ & $AP^{det}_{L}$ \\
   \shline
    Mask R-CNN(Res-50-FPN) & vanilla & 37.6 & 59.6 & 40.9 & 21.1 & 39.5 & 48.1 \\
    Mask R-CNN(Res-50-FPN) & jitter & 39.9 & 61.3 & 43.5 & 22.5 & 42.2 & 50.7  \\
    Mask R-CNN(Res-50-FPN) & map guided & \textbf{40.5}  & \textbf{62.0} & \textbf{44.2} & \textbf{23.0} & \textbf{42.7} & \textbf{51.8} \\
  \hline
    Mask R-CNN(Res-101-FPN) & vanilla & 38.2 & 60.3 & 41.7 & 20.1 & 41.1 & 50.2 \\
    Mask R-CNN(Res-101-FPN) & jitter & 42.5 & 63.7 & 46.2 & 24.3 & 45.0 & 54.2 \\
    Mask R-CNN(Res-101-FPN) & map guided & \textbf{43.0} & \textbf{64.3} & \textbf{47.2} & \textbf{24.8} & \textbf{45.9} & \textbf{54.6}  \\
  \hline
    Cascade R-CNN(Res-101-FPN) & vanilla & 43.2 & 61.6 & 47.0 & 24.1 & 46.0 & 55.4 \\
    Cascade R-CNN(Res-101-FPN) & jitter & 45.5 & 63.9 & 49.3 & 25.8 & 48.7 & 58.0 \\
    Cascade R-CNN(Res-101-FPN) & map guided & \textbf{45.9} & \textbf{64.2} & \textbf{50.0} & \textbf{26.3} & \textbf{49.0} & \textbf{58.6} \\
 \end{tabular}
 \caption{\textbf{Object detection} results on COCO test-dev, where `vanilla' denotes baseline Mask R-CNN without InstaBoost augmentation, `jitter' denotes random InstaBoost, and `map guided' denotes appearance consistency heatmap guided InstaBoost. The improvement in bounding box detection is a by-product of our InstaBoost.}
\label{tab:main_coco_det}
\vspace{-10pt}
\end{center}
\end{table*}

\begin{table*}[tb!]
\begin{center}
 \begin{tabular}{c|c|ccc|ccc}
  & Method & $AP^{seg}$ & $AP^{seg}_{50}$ & $AP^{seg}_{75}$ & $AP^{seg}_{S}$ & $AP^{seg}_{M}$ & $AP^{seg}_{L}$ \\
  \shline
    Mask R-CNN(Res-50-FPN) & vanilla & 33.8 & 56.1 & 35.5 & 14.5 & 35.3 & 49.3 \\
    Mask R-CNN(Res-50-FPN) & jitter & 35.5 & 57.9 & 37.7 & 15.7 & 37.3 & 51.6  \\
    Mask R-CNN(Res-50-FPN) & map guided & \textbf{36.0}  & \textbf{58.3} & \textbf{38.1} & \textbf{15.9} & \textbf{37.8} & \textbf{52.3} \\
  \hline
    Mask R-CNN(Res-101-FPN) & vanilla & 35.7 & 58.0 & 37.8 & 15.5 & 38.1 & 52.4 \\
    Mask R-CNN(Res-101-FPN) & jitter & 37.4 & 60.2 & 39.7 & 16.7 & 39.6 & 54.1  \\
    Mask R-CNN(Res-101-FPN) & map guided & \textbf{37.9} & \textbf{60.9} & \textbf{40.2} & \textbf{17.0} & \textbf{40.0} & \textbf{54.7} \\
  \hline
    Cascade R-CNN(Res-101-FPN) & vanilla & 37.3 & 58.8 & 40.2 & 19.4 & 40.0 & 49.8 \\
    Cascade R-CNN(Res-101-FPN) & jitter & 39.1 & 60.9 & 42.2 & 20.7 & 42.1 & 51.4 \\
    Cascade R-CNN(Res-101-FPN) & map guided &\textbf{39.5}  & \textbf{61.4}  & \textbf{42.9}  & \textbf{21.2} & \textbf{42.5}  & \textbf{52.1} \\
 \end{tabular}
 \caption{\textbf{Instance Segmentation} results on COCO test-dev, where `vanilla' denotes baseline Mask R-CNN without InstaBoost augmentation, `jitter' denotes random InstaBoost, and `map guided' denotes appearance consistency heatmap guided InstaBoost. With the help of InstaBoost, state-of-the-art instance segmentation models surpass their baseline models.}
\label{tab:main_coco_seg}
\vspace{-20pt}
\end{center}
\end{table*}

\section{Experiments}
\subsection{Datasets}
Performance of models on both bounding box detection and instance segmentation has been evaluated on popular benchmarks, including Pascal VOC~\cite{everinghamIJCV10pascal} with additional mask annotation from VOCSDS~\cite{BharathICCV11vocsds} and COCO~\cite{linECCV14coco} dataset.

\noindent \textbf{Pascal VOC and VOCSDS.} The original Pascal VOC dataset contains 17,125 images in 20 semantic categories with bounding box annotation. 2,913 images are annotated with instance masks for instance segmentation and semantic segmentation tasks. In this paper we adopted additional mask annotation from VOCSDS~\cite{BharathICCV11vocsds} with 11,355 images annotated with instance masks, following the train/test split in~\cite{liCVPR17fcis} where 5,623 images for training and 5,732 for testing.

\noindent \textbf{COCO dataset.} COCO dataset is the state-of-the-art evaluation benchmark for computer vision tasks including bounding box detection~\cite{renNIPS15fasterrcnn}, instance segmentation~\cite{liCVPR17fcis}, human pose estimation~\cite{fangICCV17rmpe} and captioning~\cite{shao2018find}. COCO is a much larger-scale image set compared to Pascal VOC, with 80 categories and more than 200,000 labeled images. Objects in COCO are annotated with both bounding box and instance mask labels. It contains large amounts of small objects, complicated object-object occlusion and noisy background, and is challenging for augmentation methods to generate ``fake'' but visually coherent images, to fully exploit the information in the dataset.

\subsection{Models}
Nowadays, Mask R-CNN~\cite{heICCV17maskrcnn} based methods are widely adopted for instance segmentation~\cite{MSCOCO} due to its promising performance and efficiency. In our experiment, we adopt the original Mask R-CNN~\cite{heICCV17maskrcnn} and its variant Cascaded Mask R-CNN~\cite{cai18cascadercnn} as our baseline networks. For Mask R-CNN, we experiment with both \textbf{Res-50-FPN} and \textbf{Res-101-FPN} backbones using open implementation~\cite{pydetectron} while only \textbf{Res-101-FPN} is tested for Cascaded Mask R-CNN based on~\cite{mmdetection2018}. Baselines are retrained using corresponding open implementations. Experimental result reveals the generalizability of our augmentation approach.

\subsection{Implementation details}
\noindent \textbf{Hyperparameters on COCO} For network training on COCO dataset, we adopt the default configuration provided by the authors, with only modifying the training epochs. We evaluated the network performance on $12$, $24$, $36$ and $48$ training epochs which are equivalent to 1x, 2x, 3x and 4x the default value in their configuration. The reported results in Tab.~\ref{tab:main_coco_det} and Tab.~\ref{tab:main_coco_seg} are obtained using $48$ training epochs. Analysis in Sec.~\ref{sec:analysis} shows that the network improves substantially after adopting our InstaBoost while suffering from over-fitting problem without such data augmentation.

\noindent \textbf{Hyperparameters on VOC} For Pascal VOC dataset, we only test the performance of \textbf{Res-50-FPN} based Mask R-CNN to evaluate the effectiveness of our algorithm. We use learning rate $5 \times 10^{-3}$ to train $20,000$ iterations, then continue training for $6,000$ iterations with $5 \times 10^{-4}$ and $4,000$ iterations with $5 \times 10^{-5}$. Other hyperparameters keep unchanged according to \textbf{Res-50-FPN} training configuration on COCO dataset.

\noindent \textbf{Hyperparameters of InstaBoost} For our random InstaBoost, we need to set the range of the uniform distribution. For the translation, the range in $x-$ and $y-$axis are set proportional  to the width and height of the object. The ratio is set as $1/15$ . For scaling, we set the range from $0.8$ to $1.2$ in our experiment. For the rotation, as described in Sec.~\ref{sec:heatmap2pos}, the degree of rotation is better small. Thus, we set the range as $[-5, 5]$. For appearance descriptor, the three fixed width contour areas are all 5 pixels and the values of weights for each contour are 0.4, 0.35 and 0.25 from inside to outside respectively. For map generation acceleration, we set the fixed size as $(180,120)$.

\subsection{Main results}
\label{sec:exp_inst_seg}
\noindent \textbf{COCO dataset} InstaBoost is evaluated with state-of-the-art instance segmentation models on the popular COCO benchmark~\cite{linECCV14coco} on both instance segmentation and bounding box detection tracks. Experimental result against competing methods in of bounding box detection is shown in Tab.~\ref{tab:main_coco_det}, and instance segmentation shown in Tab.~\ref{tab:main_coco_seg}. With InstaBoost, the performance of state-of-the-art models could be further elevated on both bounding box detection and instance segmentation tasks.

\noindent \textbf{VOC dataset} We report the instance segmentation results on VOC dataset based on R-50-FPN Mask R-CNN in Tab.~\ref{tab:main_voc}. We can see that the improvement on VOC is around $\bold{4}$ mAP, indicating the effectiveness of our method on small size dataset.

\begin{table}[tb!]
\begin{center}
\resizebox{0.48\textwidth}{!}
{
 \begin{tabular}{c|c|ccc|ccc}
  & Method & $AP^{bb}$ & $AP^{bb}_{50}$ & $AP^{bb}_{75}$ & $AP^{seg}$ & $AP^{seg}_{50}$ & $AP^{seg}_{75}$ \\
  \shline
    Mask R-CNN & vanilla &  38.06 & 68.99 & 38.06 & 38.88 & 66.18 & 40.24   \\
    Mask R-CNN & jitter & 41.76 & 71.08 & 42.37 & 42.15 & 69.06 & 44.57\\
    Mask R-CNN & map guided & \textbf{42.23} & \textbf{71.66} & \textbf{44.65} & \textbf{42.73} & \textbf{69.10} & \textbf{45.56} \\
 \end{tabular}
}
 \caption{\textbf{Object detection} and \textbf{instance segmentation} results on VOCSDS.}
\label{tab:main_voc}
\vspace{-10pt}
\end{center}
\end{table}

\begin{table}[tb!]
\begin{center}
\resizebox{0.48\textwidth}{!}
{
 \begin{tabular}{c|c|ccc}
  Translation Ratio & Scaling Ratio & $AP^{seg}$ & $AP^{seg}_{50}$ & $AP^{seg}_{75}$ \\
  \shline
    15 & 0.7-1.3 & 34.85 & 56.63 & 36.86  \\
    15 & 0.8-1.2 & \textbf{35.10} & \textbf{56.87} & \textbf{37.21} \\
    15 & 0.9-1.1 & 34.98 & 56.49 & 37.03 \\
  \hline
    1 & 0.8-1.2 & 34.51 & 55.85 & 36.74 \\
    5 & 0.8-1.2 & 34.99 & 56.51 & 37.02 \\
    15 & 0.8-1.2 & \textbf{35.10} & \textbf{56.87} & \textbf{37.21} \\
    50 & 0.8-1.2 & 35.02 & 56.59 & 37.10 \\
 \end{tabular}
}
 \caption{Sensitive analysis on different hyper-parameter configurations, on COCO val set using Res-50-FPN Mask R-CNN.}
\label{tab:sensitive}
\vspace{-20pt}
\end{center}
\end{table}

We visualize some results of Mask R-CNN trained with and w/o InstaBoost in Fig.~\ref{fig:overview}. We can see that with InstaBoost, Mask R-CNN predicts correct masks while the vanilla one generates incomplete masks or ignores the objects.

\begin{figure}[tb!]
\centering
\includegraphics[width=0.46\textwidth]{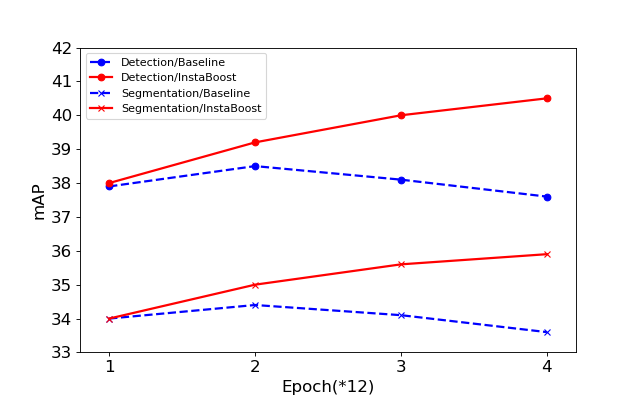}
\caption{Substantial Improvement of our data augmentation technique against overfitting.}
\label{fig:ablation_study}
\vspace{-10pt}
\end{figure}

\subsection{Analysis}
\label{sec:analysis}
\noindent \textbf{Comparison with context model}
We compare our method with previous state-of-the-art~\cite{dvornikECCV18modeling} on COCO detection and instance segmentation. We adopt Res-101-FPN Mask R-CNN as the base network. Results are given in Tab.~\ref{tab:context}. It shows that our data augmentation strategy can achieve better performance on both tasks. Moreover, ~\cite{dvornikECCV18modeling} requires extra training step and offline data prepossessing before data augmentation, while our method can be integrated into the training pipeline without tedious preparation or affecting the training efficiency.

\noindent \textbf{Comparison with random paste} To figure out the decisive role appearance consistency plays in InstaBoost, we compare our method with randomly pasting instances on the image, without overlapping with existing instances. Experiments are done on Mask R-CNN(Res-50-FPN) framework and on both VOC and COCO dataset. Tab. \ref{tab:randompaste} shows a performance degradation for 1.3 and 1.1 mAP compared to the original baseline on instance segmentation task. Such results are aligned with the findings of \cite{dvornikECCV18modeling}.

\noindent \textbf{Substantial Improvement}
We conduct experiments to validate the performance of the network using different training epochs with and without our InstaBoost. Results are shown in Fig.~\ref{fig:ablation_study}, where InstaBoost performs a promising resistance of overfitting. Both detection and segmentation accuracy of original Mask R-CNN stop increasing when epochs reaches $24$. After applying InstaBoost augmentation method, both accuracy continue going up even in large training epoch.

\noindent \textbf{Sensitivity analysis}
InstaBoost has parameters translation ratio and scaling ratio to decide the extent of the augmentation. We vary these parameters and measure AP, AP50 and AP75 of segmentation task on COCO dataset, see Tab. \ref{tab:sensitive}. For translation ratio, AP is stable in range $\frac{1}{50}$ to $\frac{1}{5}$, and drops a little when it approaches to $1$. Scaling ratio is more sensitive than translation ratio, and a variation of $0.1$ can cause about $0.1$-$0.3$ drop in AP. In our experiments, we set translation ratio to $\frac{1}{15}$ and scaling ratio to $0.8$-$1.2$.

\begin{figure}[tb!]
\centering
\includegraphics[width=0.42\textwidth]{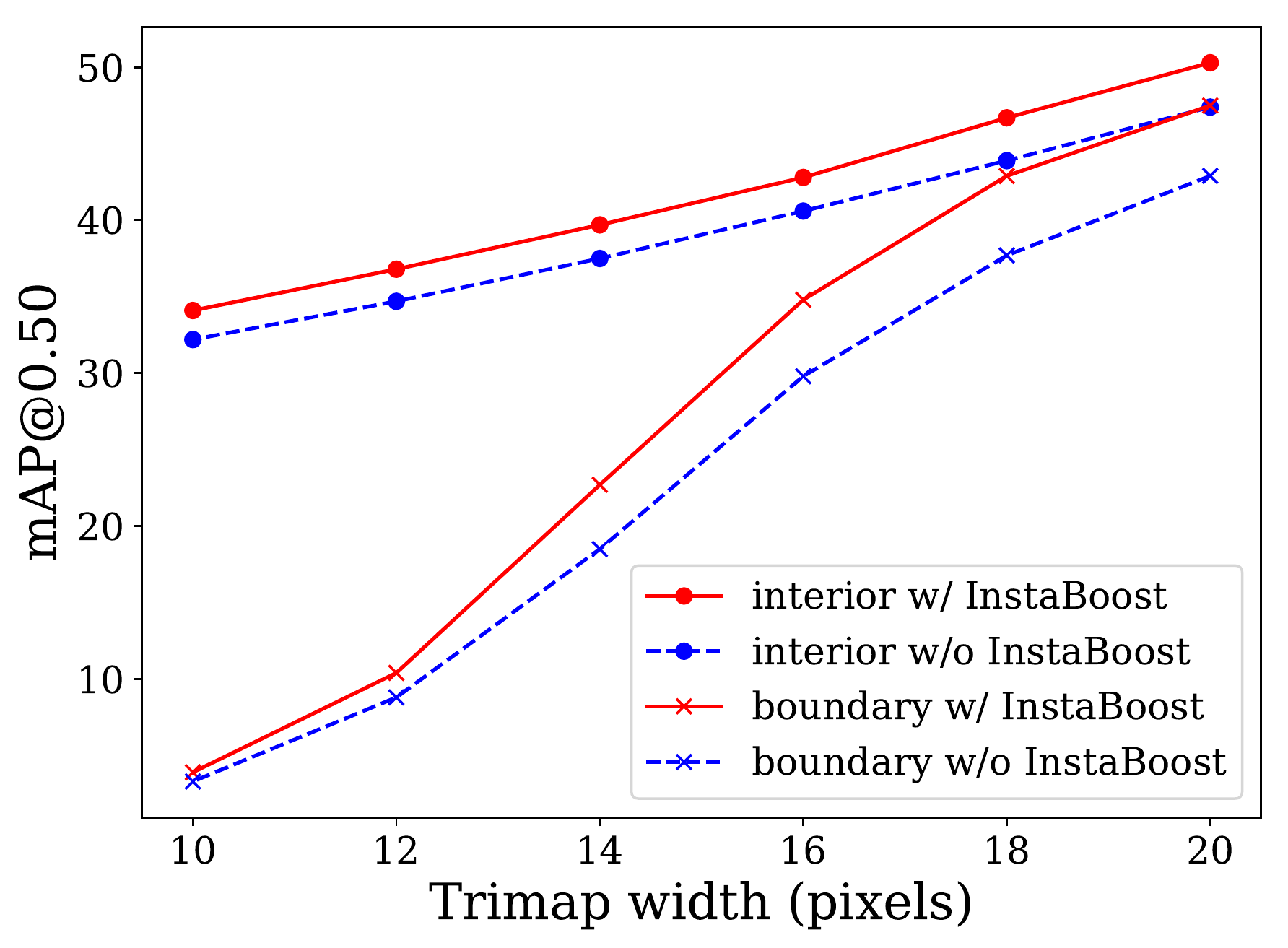}
\caption{Evaluation on interior/boundary segmentation accuracy of Mask R-CNN trained with and without InstaBoost.}
\label{fig:inter_bound}
\vspace{-10pt}
\end{figure}

\noindent \textbf{Interior-boundary study.} We compared Mask R-CNN trained with/without InstaBoost, on interior and boundary masks respectively. Following the protocol introduced in \cite{dai2015boxsup}, the interior and boundary masks are obtained from a trimap built from the edges of ground truth mask. Results in Fig.~\ref{fig:inter_bound} shows that InstaBoost improves instance segmentation accuracy on better interior detection and finer boundary prediction. The improvement on instance boundary is more significant than interior part. Readers are referred to Sec.~5.1 and Fig.~4 in \cite{dai2015boxsup} for details of this evaluation.

\begin{table}[tb!]
\begin{center}
\resizebox{0.4\textwidth}{!}
{
 \begin{tabular}{c|c|c|c}
   Method & $AP^{bb}$ & $AP^{seg}$ & Train speed(s/iter) \\
  \shline
    vanilla & 38.2 & 35.7 & 1.68  \\
    context\cite{dvornikECCV18modeling} & 38.8 & 36.2 & - \\
    ours & 43.0 & 37.9 & 1.71 \\
 \end{tabular}
}
 \caption{Comparison against context based model~\cite{dvornikECCV18modeling} on Mask R-CNN. Experimental result shows the superiority of our model in both accuracy improvement and computational overhead introduced to the running speed.}
\label{tab:context}
\vspace{-10pt}
\end{center}
\end{table}

\begin{table}[tb!]
\begin{center}
\resizebox{0.4\textwidth}{!}
{
 \begin{tabular}{c|c|c|c}
   Dataset & Method & $AP^{bb}$ & $AP^{seg}$ \\
  \shline
    VOC & vanilla & 38.06 & 38.88 \\
    VOC & random paste & 36.89 & 37.58  \\
    VOC & ours & 42.23 & 42.73 \\
    \hline
    COCO & vanilla & 37.6 & 33.8 \\
    COCO & random paste & 36.1 & 32.7 \\
    COCO & ours & 40.5 & 36.0 \\
 \end{tabular}
}
 \caption{Comparison against random paste on Mask R-CNN(Res-50-FPN). Experimental result shows appearance consistency guidance is essential.}
\label{tab:randompaste}
\vspace{-10pt}
\end{center}
\end{table}

\section{Conclusion}

This paper studies data augmentation techniques aiding the lack of training data in instance segmentation. By uniform sampling on the neighboring of identity transform in 4D transformation tuple, our simple but effective random InstaBoost achieves 1.7 mAP improvement with Mask R-CNN on COCO instance segmentation benchmark. We further devised InstaBoost with appearance consistency heatmap, reaching in total 2.2 mAP improvement on COCO instance segmentation. Our online implementation of InstaBoost can be easily embedded into existing instance segmentation frameworks, where free-lunch improvement is offered with little CPU overhead.

\section{Acknowledgement}
This work is supported in part by the National Key R\&D Program of China, No. 2017YFA0700800, National Natural Science Foundation of China under Grants 61772332.
{\small
\bibliographystyle{ieee_fullname}
\bibliography{egbib}
}

\end{document}